\documentclass[sigconf,screen]{acmart}
\AtBeginDocument{%
  \providecommand\BibTeX{{%
    \normalfont B\kern-0.5em{\scshape i\kern-0.25em b}\kern-0.8em\TeX}}}

\usepackage{tabularx}
\usepackage{makecell}
\usepackage{multirow}
\usepackage{xcolor,colortbl}

\copyrightyear{2023}
\acmYear{2023}
\setcopyright{acmlicensed}\acmConference[ICMR '23]{International Conference on Multimedia Retrieval}{June 12--15, 2023}{Thessaloniki, Greece}
\acmBooktitle{International Conference on Multimedia Retrieval (ICMR '23), June 12--15, 2023, Thessaloniki, Greece}
\acmPrice{15.00}
\acmDOI{10.1145/3591106.3592223}
\acmISBN{979-8-4007-0178-8/23/06}

\begin{document}

\title[Hypernymization for grounding]{Hypernymization of named entity-rich captions for grounding-based multi-modal pretraining}

\author{Giacomo Nebbia}
\email{gin2@pitt.edu}
\orcid{1234-5678-9012}
\affiliation{%
  \institution{University of Pittsburgh}
  \streetaddress{210 S Bouquet St}
  \city{Pittsburgh}
  \state{Pennsylvania}
  \country{USA}
  \postcode{15213}
}

\author{Adriana Kovashka}
\email{kovashka@cs.pitt.edu}
\affiliation{%
  \institution{University of Pittsburgh}
  \streetaddress{210 S Bouquet St}
  \city{Pittsburgh}
  \state{Pennsylvania}
  \country{USA}
  \postcode{15213}
}


\begin{abstract}
  Named entities are ubiquitous in text that naturally accompanies images, especially in domains such as news or Wikipedia articles. In previous work, named entities have been identified as a likely reason for low performance of image-text retrieval models pretrained on Wikipedia and evaluated on named entities-free benchmark datasets. Because they are rarely mentioned, named entities could be challenging to model. They also represent missed learning opportunities for self-supervised models: the link between named entity and object in the image may be missed by the model, but it would not be if the object were mentioned using a more common term. In this work, we investigate hypernymization as a way to deal with named entities for pretraining grounding-based multi-modal models and for fine-tuning on open-vocabulary detection. We propose two ways to perform hypernymization: (1) a ``manual'' pipeline relying on a comprehensive ontology of concepts, and (2) a ``learned'' approach where we train a language model to learn to perform hypernymization. We run experiments on data from Wikipedia and from The New York Times. We report improved pretraining performance on objects of interest following hypernymization, and we show the promise of hypernymization on open-vocabulary detection, specifically on classes not seen during training.
\end{abstract}



\begin{teaserfigure}
  \centering
  \includegraphics[width=0.5\textwidth]{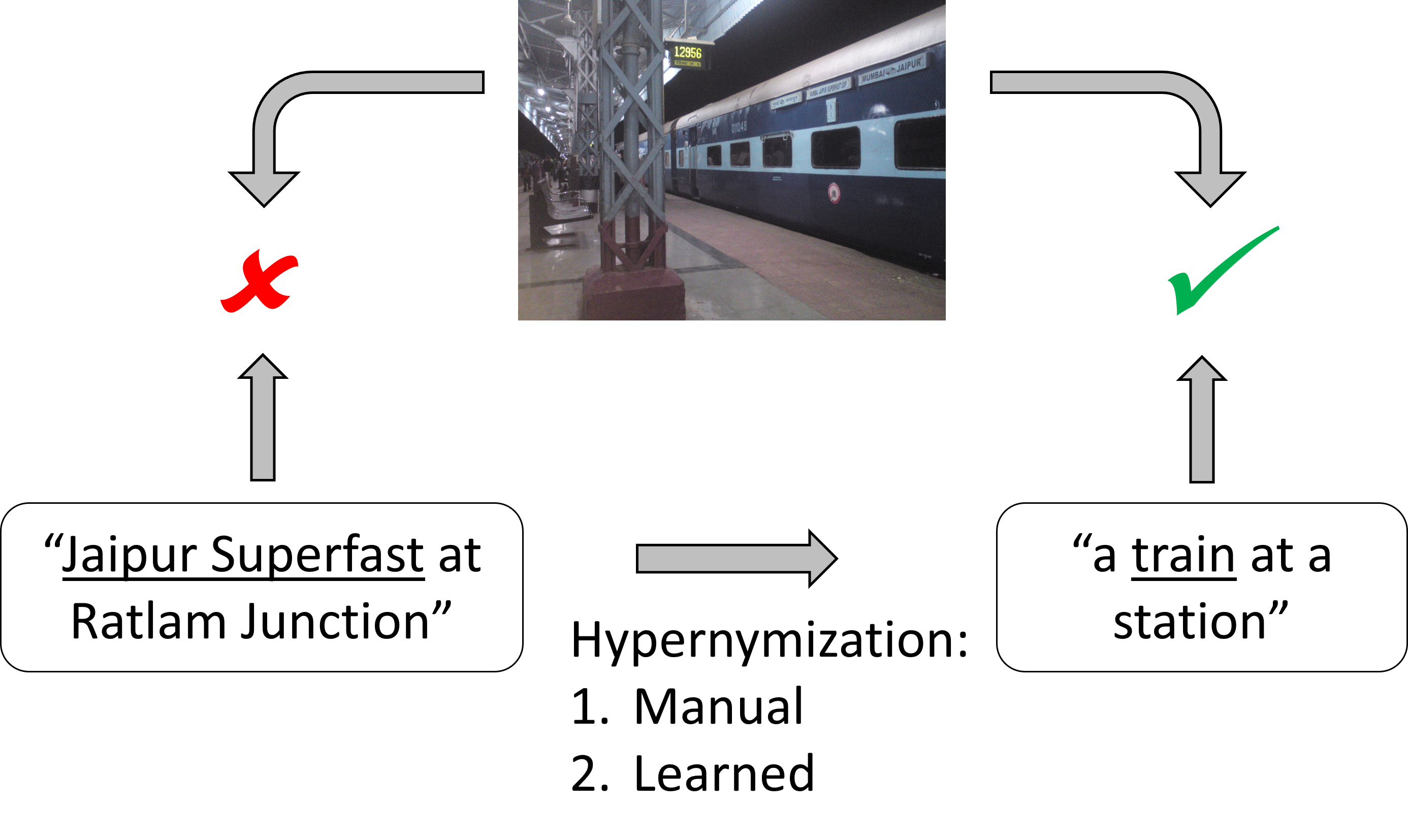}
  \caption{Our key ideas: an object mentioned using a named entity cannot be well grounded with an image. We thus introduce two methods to carry out hypernymization on the caption and show better grounding performance between the image and the hypernymized captions.}
  \Description{Our key ideas: an object mentioned using a named entity cannot be well grounded with an image. We thus introduce two methods to carry out hypernymization on the caption and show better grounding performance between the image and the hypernymized captions.}
  \label{fig:concepts}
\end{teaserfigure}


\maketitle

\section{Introduction}
\label{sec:intro}
In recent years, collections of large numbers of image-caption pairs~\cite{Changpinyo2021,Sharma2018,srinivasan2021wit} have made training large  (i.e., hundreds of millions of parameters), general-purpose computer vision models~\cite{yuan2021florence,radford2021learning,jia2021scaling} possible. Such models can later be used as building blocks for or fine-tuned on tasks of interest~\cite{shi2022proposalclip,zhong2022regionclip}. The captions in these large image-text datasets are not manually collected, but ``scraped'' from existing sources as they naturally accompany the images (e.g., the \verb+alt-text+ HTML field for images crawled from the Internet~\cite{Changpinyo2021,Sharma2018}, or text describing images from Wikipedia~\cite{srinivasan2021wit} or news articles~\cite{Tran_2020_CVPR,biten2019good}). However, the use of captions naturally associated with images presents some challenges: captions can be ill-formed or irrelevant~\cite{Sharma2018}, and they may include named entities (NEs)~\cite{Sharma2018}. 

NEs represent a challenge when pretraining multi-modal models because they are rarely mentioned, making it hard for a model to learn the link between the NE and the corresponding object in the image. In fact, previous work~\cite{srinivasan2021wit} pointed at NEs as the reason for sub-par performance of multi-modal retrieval models trained on NE-rich Wikipedia data and tested on NE-free COCO~\cite{lin2014microsoft} and Flickr30K~\cite{young2014image}.

Some domains contain a large fraction of captions with NEs, and these NEs require special handling. 
Captions gathered from the \verb+alt-text+ HTML field may include few NEs (e.g., only around 25\% of captions in CC12M~\cite{Changpinyo2021} originally included a NE for a person), but the majority of captions in domains like Wikipedia and news articles include NEs (e.g., more than 95\% in some news datasets~\cite{Tran_2020_CVPR,biten2019good}). 
Thus, \emph{discarding} captions with NEs is not feasible. Furthermore, we argue that \emph{ignoring} NEs (as often done when pretraining multi-modal models~\cite{yuan2021florence,jia2021scaling}) represents a missed learning opportunity: had the object been referenced with its name rather than with a NE, the image-caption pair could have been used to learn a better representation for the object. 
Following this reasoning, we propose to address NEs through hypernymization: replacing a NE with its hypernym (i.e., a general term representing the class/category of the NE). Our key idea is summarized in Figure~\ref{fig:concepts}.

We investigate how hypernymization can be used on captions with named entities to improve grounding-based pretraining and open-vocabulary object detection. Grounding-based pretraining aims to match image regions with their corresponding word tokens from the caption. Because of this fine-grained matching, we hypothesize that training of such models may particularly be impacted by the presence of NEs. In addition, we fine-tune on open-vocabulary detection, which refers to zero-shot detection~\cite{bansal2018zero} with captions used as a source of supervision. Since the goal of grounding is to learn representations for objects mentioned in the captions, open-vocabulary detection allows us to evaluate their quality for objects the model is fine-tuned on as well as for objects the model is not fine-tuned on.

We introduce two methods to hypernymize captions: 
\begin{enumerate}
    \item ``manual hypernymization'', where we apply a pipeline relying on named entity recognition and on a comprehensive ontology of concepts
    \item ``learned hypernymization'', where we train a language model to perform hypernymization based on the context surrounding NEs in a caption
\end{enumerate}

We apply our proposed methods to captions from Wikipedia and from The New York Times and we compare pretraining on the original, NE-rich captions with pretraining on their hypernymized counterparts. After pretraining, we fine-tune models on open-vocabulary object detection and analyze how improvements in pre-training performance translate to improvements in downstream performance. We report improved pretraining performance on subsets of classes of interest and we highlight the challenges related to hypernymization.

The rest of the manuscript is organized as follows: Section~\ref{sec:related_work} covers related work, Section~\ref{sec:methods} introduces our proposed hypernymization approaches, Section~\ref{sec:experiments} describes our experimental design, Section~\ref{sec:results} reports our results, and Section~\ref{sec:discussion} presents a discussion of our results and our conclusions.

\begin{table*}[htbp]
    \centering
    \begin{tabular}{l|l|l|l}
        \textbf{Named Entity} & \textbf{Hypernyms} & \textbf{Most Specific} & \textbf{Lowest Common}\\
        \hline \hline
         \multirow{2}{*}{Class 319/4} & Train / MeanOfTransport & \multirow{2}{*}{Train} & \multirow{2}{*}{MeanOfTransport} \\
         & MeanOfTransport & & \\
         \hline
         \multirow{4}{*}{Curtly Ambrose} & Person / [\ldots] & \multirow{4}{*}{Cricketer} & \multirow{4}{*}{Thing} \\
         & Athlete / Person / [\ldots] & & \\
         & Cricketer / Athlete / [\ldots] & & \\
         & Agent & &
    \end{tabular}
    \caption{Comparison between ``most specific'' and ``lowest common ancestor'' methods to select among multiple DBPedia Ontology types returned for a given NE. The slashes indicate the path to the root of the ontology (i.e., Thing, omitted for brevity). For brevity, we omit full paths to the root when needed.}
    \label{tab:ne:choose_hypernym}
\end{table*}

\section{Related Work}
\label{sec:related_work}
\textbf{Self-supervision from captions.} A common approach in current self-supervision for computer vision is to take advantage of the naturally co-occurring captions associated with images crawled from the web as a source of ``free'' supervision~\cite{radford2021learning,yuan2021florence,lu2019vilbert}. The main way to leverage this source of supervision is Image-Text Matching, which trains models to distinguish between matching image-text inputs (i.e., those images and captions paired in the dataset) and non-matching ones (i.e., any image and caption not paired with each other)~\cite{lu2019vilbert,li2019visualbert,huang2020pixel,radford2021learning,jia2021scaling,chen2020uniter}. This idea has also been extended to image regions and word tokens: parts of an image and parts of its caption are matched with each other, a task known as grounding~\cite{Zareian2021,gu2022vild,nebbia2022doubling}.

Due to the success of multi-modal pretraining, interest in image-caption datasets has grown, and so has the size of these datasets: from 3/12 million in Conceptual Captions (CC)~\cite{Changpinyo2021,Sharma2018} to 400 and 900 million in CLIP~\cite{radford2021learning} and Florence~\cite{yuan2021florence}, respectively. 
With datasets of such magnitude, manual inspection of the text is not feasible, so quality checks must be implemented during~\cite{radford2021learning} or after~\cite{Changpinyo2021,Sharma2018} collection. For example, CLIP collected captions so that they would include common words as found in Wikipedia to ensure a broad variety of visual concepts was covered, while CC removed all captions with high rate of token repetition.
\\
\textbf{Named entities.} Among the potential problems with multi-modal datasets, previous work has specifically highlighted named entities as an issue of interest for multi-modal supervision~\cite{Sharma2018,srinivasan2021wit}. In some datasets, this problem may not be pervasive~\cite{Changpinyo2021}, and models can be trained without addressing it~\cite{jia2021scaling,yuan2021florence,radford2021learning}. In other domains, though, NEs are dominant~\cite{Tran_2020_CVPR,biten2019good} and simply ignoring them~\cite{srinivasan2021wit} has shown to lead to underperforming models. Few studies used hypernymization as a pre-processing steps~\cite{Changpinyo2021,Sharma2018}, but not for NE-rich domains. We address this gap in the literature by investigating the issue posed by NEs while pretraining multi-modal models in NE-rich domains, and evaluating the impact on grounding-based pretraining and on downstream object detection.
\\
\textbf{Pretraining evaluation.} Evaluation of self-supervised models is an active research direction. A common approach is to fine-tune on many downstream tasks of interest~\cite{radford2021learning}, assuming that better pretraining equals better downstream performance. Testing on such a variety of downstream tasks also assumes that better feature representations (and thus better pretraining strategies) are generalizable to many tasks of interest. Recent work~\cite{wei2021aligning,roh2021spatially,zhong2021dap} has started challenging this view by suggesting that pretraining should be tailored toward a specific task of interest. In particular, initial evidence from the vision literature~\cite{ericsson2021well} shows that, at the current state, no single pretraining strategy outperforms all others regardless of downstream task.
We contribute to the research on how to evaluate pretrained models by adapting a previous study~\cite{gao2022open} to this task. In addition, we follow the idea of coupling pretraining and finetuning~\cite{wei2021aligning,roh2021spatially,zhong2021dap} by choosing a downstream task closely related to grounding: open-vocabulary object detection.
\\
\textbf{Object detection.} Object detection is a benchmark downstream task that is closely related to grounding pretraining. In particular, previous work~\cite{Zareian2021,nebbia2022doubling} has fine-tuned grounding models on open-vocabulary object detection~\cite{zhong2022regionclip,du2022learning}, where no samples of some classes are available during training (like in zero-shot detection~\cite{bansal2018zero}), but captions are available to provide supervision in the pretraining stage. We follow previous studies and evaluate the effect of hypernymization after pretrained models are fine-tuned for open-vocabulary object detection.

\section{Methods}
 \label{sec:methods}
In this section, we detail our proposed hypernymization strategies and we provide a summary of the grounding pretraining architecture we use~\cite{Zareian2021}.

\subsection{Manual Hypernymization}
\label{sec3_manual}
For our first hypernymization approach, we rely on a named entity recognition (NER) system and on a comprehensive knowledge base where we can look up each NE. We call this approach ``manual'' as it mimics how a person would carry out hypernymization.

The main strength of this approach is the use of a NER system and of a knowledge base, which makes for a very competitive hypernymization method; if we removed such resources, we would sacrifice very informative tools.

The first step in our manual hypernymization pipeline is NER, where NEs are identified within each caption (e.g., ``Class 319/4'' from caption ``The first refurbished Class 319/4''). Generally, NER algorithms return a label for each NE, but the domain for these labels is limited~\cite{akbik2019flair,finkel2005incorporating,bird2009natural}  (e.g., Person, Location, Organization, and Miscellaneous). For this reason, we look up each NE on DBPedia~\cite{lehmann2015dbpedia}, a semantic network of concepts extracted from Wikipedia. DBPedia itself matches each query NE to a list of its entities~\footnote{https://www.dbpedia.org/resources/lookup/, last accessed April 5th, 2023}, from which we select the highest scoring one (the scoring is implemented by DBPedia). The selected entity is associated with multiple ``types'', defined in the DBPedia Ontology (e.g., ``Class 319/4'' is associated with ``Train'' and ``Mean Of Transport''). We pick the most specific type, defined as the farthest from the root of the DBPedia Ontology (e.g., ``Train'', which is a child of ``Mean Of Transport'' - see Table~\ref{tab:ne:choose_hypernym}). Alternative approaches to selecting one type include: (a) the closest category of interest, and (b) the lowest common ancestor (i.e., among all ancestors shared by the returned types, the one that is farthest from the root). We discard the first one because we want to keep the method independent of any list of pre-defined objects. To decide between the ``most specific type'' approach and the ``lowest common ancestor'' approach, we evaluate some examples (like those in Table~\ref{tab:ne:choose_hypernym}), and select the ``most specific type'' alternative.

Finally, if a NE is not found in DBPedia, we remove it. Given that our motivation is that NEs are hard for models to ground, we aim to leave no NEs in the captions.

While touted as a strength of our proposed manual hypernymization approach, relying on a NER tool and on a knowledge base is also a weakness: for example, if a NE is missed, it would be impossible to hypernymize, and if only partially recognized, the NE may not be found in the knowledge base. In addition, if hypernymization relies on a knowledge base look-up operation, the knowledge base must include all possible NEs and must be kept up to date constantly. Since these requirements are very restrictive, we next propose a method that relaxes them.

\subsection{Learned Hypernymization}
\label{sec::methods:learned}
Our second hypernymization approach is to train a language model to perform hypernymization. With this approach, we aim to relax the two constraints introduced by the manual approach (i.e., a NER system and an all-encompassing knowledge base). With this new approach, we still rely on a list of NEs and their hypernyms, but we do not require such a list to be exhaustive. In addition, we want to remove the issue of propagating errors from NER to hypernymization by merging the two steps.  

\begin{figure*}[htbp]
    \centering
    \includegraphics[width=0.8\linewidth]{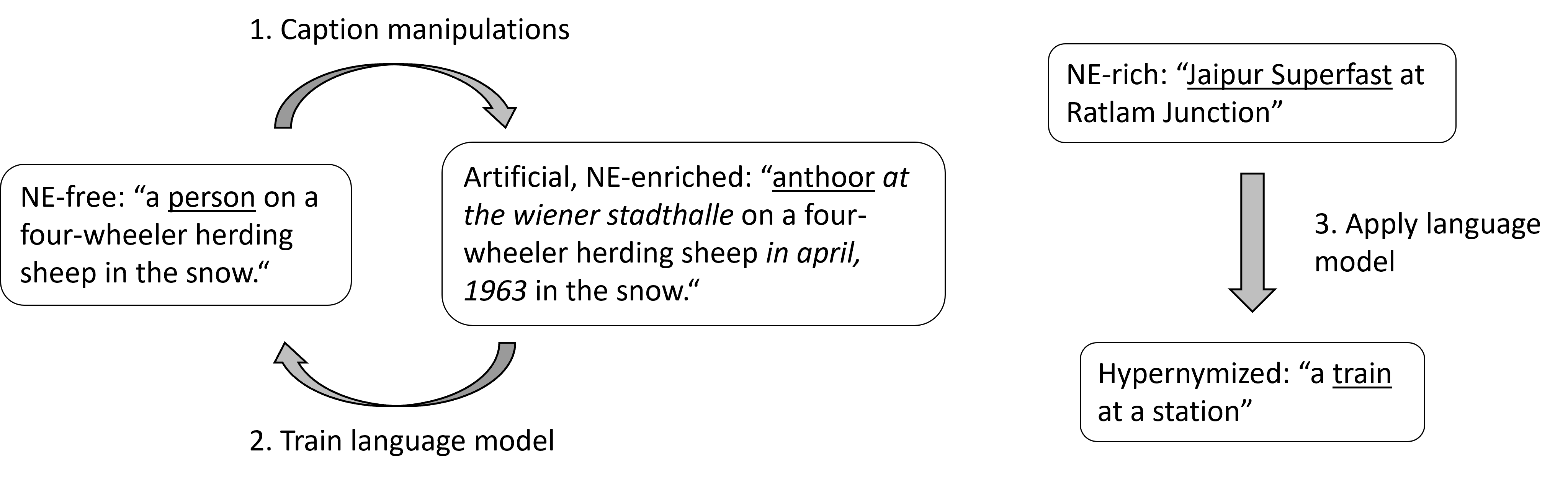}
    \caption{Proposed pipeline for learned hypernymization. 1) We apply pre-defined manipulations to NE-free captions to make them more similar to captions from NE-rich settings. 2) We train a language model to reconstruct the original, NE-free captions from their artificial, NE-enriched versions. 3) We apply the trained language model to NE-rich captions to perform hypernymization.}
    \label{fig:learned_hypern:summary}
\end{figure*}

It is not straightforward to train a language model for the hypernymization task in a supervised way because we do not have ground truth caption pairs with (NE-rich, hypernymized) caption pairs. 
Instead of paired data, we separately have (1) NE-rich captions (e.g., Wikipedia~\cite{srinivasan2021wit} and news articles~\cite{Tran_2020_CVPR}), and (2) NE-free captions (e.g., COCO~\cite{chen2015microsoft}). 
We thus create an artificial, NE-enriched version of the NE-free captions by including NEs and other characteristics typical of NE-rich settings. We then train a language model to reconstruct the original, NE-free caption from the NE-enriched captions. The hypothesis underlying this approach is that captions contain enough information to learn how to hypernymize NEs. After training this model, we apply it to a NE-rich setting and generate hypernymized captions. Figure~\ref{fig:learned_hypern:summary} illustrates this process with examples.

We next describe the learned hypernymization pipeline in more detail. The first step in creating artificial, NE-enriched captions is to curate a list of NEs that can be introduced in the original, NE-free captions. To do so, we use DBPedia and retrieve NEs for each ``type'' (i.e., hypernym) included in the DBPedia Ontology. While this step requires the use of DBPedia, we argue that its use in the learned approach is less restrictive than its use in the manual one. In fact, the manual approach requires DBPedia to include all possible NEs, while the learned approach only requires it to include enough NEs to train a model to perform hypernymization.  

With these lists available, we apply the following manipulations:
\begin{enumerate}
    \item We replace mentions of DBPedia Ontology types with a random NE from the corresponding list with probability $p_{NE}$. We add more than one NE if a type is mentioned in plural form (e.g., ``a group of $<$type$>$''). This step is crucial to teach the language model to perform hypernymization.
    \item We randomly add locations (and dates) at the beginning and end of a caption with probability $p_{date-loc}$ and in the middle of a caption with probability $p_{middle}$. To make sure the captions remain well-formed sentences, we add locations (and dates) in the middle of a caption only before punctuation or prepositions. Dates and locations are often found in NE-rich datasets, and we want the model to remove them as they do not carry information relevant to grounding (or detection) and could create spurious grounding relationships (e.g., if captions mentioning ``Los Angeles'' show boats, the model may learn to ground the two).
    \item We add artificial captions with NEs only that are matched to empty sentences. We add these captions to teach the model to remove (and not randomly hypernymize) NEs if context is insufficient. Incorrect hypernymization would, in fact, create spurious grounding relationships.
    \item  We include captions with no mention of objects of interest. For these captions, we do not replace mentions of DBPedia Ontology types with NEs, but we add dates and locations. The goal is to teach the model not to add object mentions in every caption. Hallucinating objects in captions would create a sample that confuses the model since it would establish a grounding relationship that does not exist.
\end{enumerate}

\subsection{Grounding Architecture}
\label{sec:methods:grounding}
We use OVR-CNN~\cite{Zareian2021} as our grounding model, which consists of a visual backbone and a text encoder whose outputs are combined through self-attention. The visual backbone extracts features for each element of a grid defined over the input image and passes them through a visual-to-language (V2L) layer that maps the visual embedding space to the text embedding space. The input caption is processed by the text encoder and a multi-layer transformer combines the text and visual features into output features on which the loss functions are defined. In addition to the standard Image-Text Matching (ITM) loss~\cite{lu2019vilbert,li2019visualbert,huang2020pixel,radford2021learning,jia2021scaling,chen2020uniter} and Masked Language Model (MLM) loss~\cite{chen2020uniter,lu2019vilbert,su2019vl}, OVR-CNN pretrains models using a grounding loss defined as follows.

Let $e^C_j$ be the text embedding for the $j$-th caption token, $n_C$ the number of tokens, $e^I_i$ the V2L embedding for the $i$-th image region, and $n_I$ the number of regions. Grounding $\langle I,C \rangle_G$ between image $I$ and caption $C$ is defined as 
\begin{equation}
    \label{eq:grounding}
    \langle I,C \rangle_G = \frac{1}{n_C}\sum_{j=1}^{n_C}\sum_{i=1}^{n_I} a_{i,j}\langle e_i^I, e_j^C \rangle_L
\end{equation}
with $\langle e_i^I, e_j^C \rangle_L$ the dot product between $e_i^I$ and $ e_j^C$, and the coefficient $a_{i,j}$ defined as
\begin{equation}
    \label{eq:alpha}
    a_{i,j} = \frac{\exp \langle e_i^I, e_j^C \rangle_L}{\sum_{i'=1}^{n_I} \exp \langle e_{i'}^I, e_j^C \rangle_L}
\end{equation}
The attention coefficients $a_{i,j}$ are computed as the softmax of each embedding pair's dot product across image regions, and they are used as weights to average the region-token embedding dot product in Equation~\ref{eq:grounding}.

The grounding loss should encourage overall grounding for an image and its matching text to be maximized, while grounding between each image (caption) and a non-matching caption (image) should be minimized. Two grounding losses are introduced, where, given an image-caption pair, all other captions in the batch are used as negative examples for the image, and all other images are used as negative examples for the caption. These two grounding losses are, respectively
\begin{equation}
    \label{eq:l_I}
    L_G(I) = -\log \frac{\exp \langle I,C \rangle_G}{\sum_{C'' \in B_C} \exp \langle I, C'' \rangle_G}
\end{equation}
and
\begin{equation}
    \label{eq:l_C}
    L_G(C) = -\log \frac{\exp \langle I,C \rangle_G}{\sum_{I' \in B_I} \exp \langle I', C \rangle_G}
\end{equation}
The final loss is the sum of the two grounding losses and the ITM and MLM losses.
\begin{equation}
    L(I,C) = L_G(I)+ L_G(C) + L_{ITM} + L_{MLM}
\end{equation}

\section{Experimental Design}
\label{sec:experiments}

\subsection{Datasets}
\label{sec:experiments:datasets}
We analyze hypernymization on two NE-rich datasets: Wikipedia Image-Text (WIT)~\cite{srinivasan2021wit} and NYTimes800k~\cite{Tran_2020_CVPR}.  

WIT includes images and text extracted from Wikipedia. Each image is associated with multiple sources of text, some of which may be in multiple languages. In detail, each image can be associated with: (1) a reference description (i.e., the caption visible on the Wikipedia page), (2) an attribution description (i.e., the text found on the Wikimedia page of the image), or (3) the \verb_alt-text_ description (i.e., the HTML field associated with the image). We concatenate the reference text and the English-only part of the attribution description to create a caption for each image, as done by~\cite{srinivasan2021wit}. Because of the dataset size, we subset the data by excluding empty captions as well as all images not in jpeg format (since gif, png, and svg files are likely to be graphics and not photographs), greyscale images (likely old photographs), and images whose captions mention dates before 1950 (likely scans of old photographs). This results in 303,589 image-caption pairs. We hold out a validation set (N=18,755) for hyperparameter tuning and model selection.

NYTimes800k~\cite{Tran_2020_CVPR} includes 445K articles and 793K images with captions from The New York Times spanning 14 years, and it was collected using The New York Times API. It follows a similar collection pipeline as GoodNews~\cite{biten2019good}, but it is 70\% larger and more complete (GoodNews includes some incomplete articles and some non-English text). Named entities are dominant in this dataset: 97\% of captions include at least one~\cite{Tran_2020_CVPR}. We use the official training and validation splits for model pre-training and hyperparameter tuning and model selection.

For fine-tuning on open-vocabulary detection, we use MS-COCO Objects~\cite{lin2014microsoft}, which includes 118,287 training images and 5,000 validation images. We test models on the official validation set, and we hold out a subset of 5,000 training images as an internal validation set for hyperparameter tuning and model selection.

We train our hypernymization language model on MS-COCO Captions~\cite{chen2015microsoft} and its NE-enriched version (Section~\ref{sec::methods:learned}). MS-COCO Captions includes an average of 5 captions for each MS-COCO Objects image. We hold out the same subset of the training split for internal validation.

Finally, we include captions from Conceptual Captions (CC)~\cite{sharma2018conceptual} that do not mention any COCO object when training our learned hypernymization approach. CC includes 3M image-text pairs from the Internet, where captions are pre-processed versions of the \verb+alt-text+ field associated with each image.

\subsection{Baselines}
\label{sec:experiments:baselines}
As our baseline hypernymization strategies, we use two simple ways to deal with named entities: ignore them by not modifying the captions (as often done in the literature~\cite{yuan2021florence,jia2021scaling}) or remove them. 

For grounding pre-training, we train models on these two baseline versions of the captions and compare them with models pre-trained on captions hypernymized using our two proposed methods.

For open-vocabulary detection, we fine-tune the grounding model pre-trained on the original captions for WIT (and NYTimes800k) and use it as our baseline. We compare this baseline to models fine-tuned from grounding models pre-trained on captions hypernymized with our proposed approaches.

\subsection{Language Model Evaluation}
After training a language model to perform hypernymization (Section~\ref{sec::methods:learned}), we start by verifying that it is able to reconstruct the original, NE-free captions. To do so, we run the learned model on the validation split of COCO captions and compute Rouge~\cite{lin2003automatic} metrics to quantify the overlap between original and reconstructed COCO captions. To verify the need to fine-tune the language model on our NE-enriched captions, we evaluate an off-the-shelf, non-fine-tuned version of the same model and use it as a baseline.

\subsection{Hypernymized Datasets Evaluation}
After verifying that the learned language model can reconstruct COCO captions, we apply it to our NE-rich datasets and extract dataset statistics for the original captions and their two hypernymized versions. Specifically, we extract number of unigrams and average length of caption (as number of unigrams in the caption). The purpose of these statistics is to numerically compare the three versions of the NE-rich datasets and verify that our manipulations are having the desired effect of making captions more similar to those in a NE-free domain.

Following~\cite{srinivasan2021wit}, we compute the Jensen-Shannon Divergence (JSD) between the unigram distribution of COCO (train) and that of the three NE-rich dataset versions (original and the two hypernymized). JSD values are low if the distributions are similar, so we aim to show how hypernymization transforms NE-rich settings to be more similar to COCO and thus to be better suited for training models for grounding and object detection.

\subsection{Object Mention Extraction}
\label{sec:experiments:mentions}
For our grounding evaluation (to follow in Section~\ref{sec:experiments:grounding}) we ground images with mentions of COCO classes in their captions. To extract such mentions, we use \verb+ExactMatch+~\cite{unal2022learning}: only verbatim occurrences of COCO classes are counted as mentions. We evaluate the ability of \verb+ExactMatch+ to correctly extract object mentions from COCO captions by computing precision and recall with respect to the ground truth labels provided for each image. In detail, we extract the set of unique objects mentioned by all captions describing an image and we compare them with the unique set of ground truth labels associated with the same image.

This evaluation allows us to identify classes for which we can reliably extract grounding maps: if precision is high for a class, we can expect an object mention to correspond to an actual object in the image.

\subsection{Grounding Evaluation}
\label{sec:experiments:grounding}
To evaluate grounding models, we adapt a strategy previously introduced to assign pseudo ground truth labels to region proposals~\cite{gao2022open}. Given a caption and an image, we compute a grounding map between each mention of a COCO class (Section~\ref{sec:experiments:mentions}) and the image (i.e., the attention coefficients defined by Equation~\ref{eq:alpha}). We then compute the average grounding coefficient within each ground truth bounding box for the image and select the box with highest average as the chosen detection result. Once bounding boxes and their scores are computed for each (image, caption) pair, we carry out a standard detection evaluation and report mAP across the 80 COCO classes. The use of ground truth bounding boxes allows us to evaluate the potential of the learned grounding for object detection. Given that grounding is a preliminary step, we would rather have a model that correctly grounds most pixels within ground truth bounding boxes (at the cost of potentially more incorrect grounding outside of them). This would mean that the model learns a (crude) representation for, at the least, the objects of interest. Further refinement of such representations can be obtained with fine-tuning.
We only compare detection results to ground truth bounding boxes for classes mentioned by the captions.

\subsubsection{Fine-grained grounding evaluation.} Because DBPedia is a general-purpose tool not designed for a specific task or dataset, the set of concepts it represents does not coincide with that represented by COCO classes. We thus focus on classes included in the DBPedia Ontology, where we specifically expect hypernymization to be beneficial. Both the manual and learned methods can directly improve performance on these classes since they can look them up in DBPedia and perform hypernymization or obtain training data, respectively. 
Our learned approach 
could still infer how to hypernimize classes not in the ontology by leveraging context in the caption (e.g., by learning the context around instances of ``bicyle'', our language model may learn to hypernymize NEs of that type without needing artificial captions where mentions of ``bicycle'' are replaced with NEs). 

Finally, because our grounding evaluation depends on extraction of object mentions, we focus on mentions that very likely refer to the correct object by excluding classes for which \verb|ExactMatch|'s precision is lower than the overall average. 

\subsection{Open-Vocabulary Detection Evaluation}
\label{sec:experiments:ovr}
To further evaluate the impact of hypernymization on quality of pretraining, we fine-tune the pretrained model on the open-vocabulary detection task~\cite{Zareian2021}, where only a subset of 48 ``base'' classes are seen during training, while the model is also tested on 17 ``target'' classes~\cite{bansal2018zero}. Following~\cite{Zareian2021}, we report performance as mAP@0.5.

\subsection{Implementation}
\label{sec:experiments:implementation}
To perform named entity recognition, we use the off-the-shelf Flair tagger~\cite{akbik2019flair}, available from Hugging Face~\cite{wolf-etal-2020-transformers}.

For our learned hypernymization approach, we use the T5-small~\cite{raffel2020exploring} language model, which we fine-tune for 5 epochs on one Google Cloud TPU with learning rate of 0.0001, batch size of 12, and gradient accumulation step of 8. Given the affinity of caption reconstruction with summarization, we add the ``summarize:'' prompt to the beginning of each input caption.

To create the artificial, NE-enriched captions, we set $p_{NE}=0.7$, $p_{date-loc}=p_{middle}=0.3$ (Section~\ref{sec::methods:learned}).

For pre-training, we adapt the code from~\cite{Zareian2021} with a ResNet50~\cite{he2016deep} as the visual encoder and a frozen BERT-base model~\cite{devlin-etal-2019-bert,wolf-etal-2020-transformers} as the text encoder. We set the learning rate to 0.001 (decreased by a factor of 10 at 50\% and 80\% of training). We set the batch size to 9 and we train using 3 Nvidia Titan X GPUs. We select hyperparameters after a grid search on our held-out validation set.

For open vocabulary detection, we freeze the first two layers of the ResNet50 backbone and fine-tune the rest with learning rate of 0.005 (decreased as before) and batch size of 8 on 2 Nvidia Titan X GPUs.

\section{Results}
\label{sec:results}

\subsection{Baselines}
\label{sec:results:baselines}
Table~\ref{table:results:baselines} reports pretraining evaluation results for a model trained on COCO and for our baseline models trained on WIT and NYTimes800k.
\begin{table}[htbp]
    \centering
    \begin{tabular}{l|c}
        \textbf{Dataset} & \textbf{mAP}\\
         \hline \hline
        COCO & 54.3  \\
        \hline \hline
        WIT & 38.5 \\
        \hline
        WIT - no NEs & 37.5 \\
        \hline \hline
        NYTimes800k & 40.5 \\
        \hline
        NYTimes800k - no NEs & 40.6
    \end{tabular}
    \caption{Pretraining evaluation performance (mAP, in percentage) on MS-COCO val 2017 for models pretrained on COCO, WIT (with and without NEs) and NYTimes800k (with and without NEs)}
    \label{table:results:baselines}
\end{table}

The model pretrained on COCO represents an upper bound for our experiments since it is trained and evaluated on the same dataset. We observe a performance gap between this upper bound and models pretrained on WIT or NYTimes800k. This is due to reasons including domain shift and differences in captions' characteristics (e.g., WIT's captions tend to be more narrative and redundant while COCO's are more descriptive and succinct). In addition, we notice how our two baseline approaches (i.e., original captions and removing NEs) perform on par with each other for both WIT and NYTimes800k, indicating that the models seem to be able to largely ignore named entities. This is an interesting finding: NEs do not significantly contribute to grounding-based pretraining; in a way, they are ``wasted''. This motivates investigating how to better leverage the supervision NEs could provide.

\subsection{How well does a language model perform hypernymization?}
\label{sec:results:lm_eval}
\begin{table}[htbp]
    \centering
    \begin{tabular}{l*{3}{|c}}
        \textbf{Fine-tuned} & \textbf{Rouge1} & \textbf{Rouge2} & \textbf{RougeL} \\
        \hline \hline
        No & 59.50 & 47.39 & 58.81 \\
        \hline
        Yes & 91.49 & 89.36 & 91.48 
    \end{tabular}
    \caption{Rouge metrics evaluating the ability of the learned language model to reconstruct COCO captions from their artificial, NE-enriched versions.}
    \label{table:results:lm_eval}
\end{table}
Table~\ref{table:results:lm_eval} reports the evaluation of the language model trained to perform hypernymization. We notice how fine-tuning is necessary since an off-the-shelf, non-fine-tuned model is not able to reconstruct the original COCO captions, while a fine-tuned version of the model achieves high reconstruction performance.

\subsection{How similar are hypernymized captions to COCO captions?}
\label{sec:results:datasets_stats}
Table~\ref{table:results:datasets_stats_wit} reports unigram-level statistics for COCO, WIT, and its two hypernymized versions, while Table~\ref{table:results:datasets_stats_nyt} shows them for NYTimes800k and its hypernymized versions.
\begin{table}[htbp]
    \centering
    \begin{tabular}{l|l|l|l|l}
         & \textbf{WIT} & \textbf{\makecell[l]{WIT\\manual}} & \textbf{\makecell[l]{WIT\\learned}} & \textbf{COCO}\\
         \hline \hline
         Words & 366,223 & 162,327 & 162,254 & 29,650\\
         \hline
         Length & 24 & 21 & 21 & 11
    \end{tabular}
    \caption{Dataset statistics for COCO and WIT and its two hypernymized versions. Length represents the average number of unigrams per caption.}
    \label{table:results:datasets_stats_wit}
\end{table}

\begin{table}[htbp]
    \centering
    \begin{tabular}{l|l|l|l|l}
         & \textbf{NYT} & \textbf{\makecell[l]{NYT\\manual}} & \textbf{\makecell[l]{NYT\\learned}} & \textbf{COCO}\\
         \hline \hline
         Words & 224,048 & 57,471 & 120,264 & 29,650\\
         \hline
         Length & 23 & 21 & 21 & 11
    \end{tabular}
    \caption{Dataset statistics for COCO and NYTimes800k (NYT) and its two hypernymized versions. Length represents the average number of unigrams per caption.}
    \label{table:results:datasets_stats_nyt}
\end{table}
From Tables~\ref{table:results:datasets_stats_wit} and~\ref{table:results:datasets_stats_nyt}, we notice how both hypernymization strategies reduce the number of unique unigrams and the average length of a caption, moving these statistics closer to their values for COCO.

To further evaluate how hypernymization shifts the unigram distribution of NE-rich captions toward that of COCO captions, Table~\ref{table:results:jsd} reports the Jensen-Shannon Divergence (JSD) between the unigram distribution for COCO captions and different versions of WIT and NYTimes800k captions.
\begin{table}[htbp]
    \centering
    \begin{tabular}{l|c|c}
        \textbf{Dataset} & \textbf{COCO v. WIT} & \textbf{COCO v. NYTimes800k}\\
        \hline \hline
        Original & 0.597 & 0.557 \\
        \hline
        Manual & 0.584 & 0.552 \\
        \hline
        Learned & 0.529 & 0.501
    \end{tabular}
    \caption{Jensen-Shannon Divergence (JSD) between datasets. Low values indicate similar distributions.}
    \label{table:results:jsd}
\end{table}

We observe how hypernymization is successful in moving the unigram distribution of WIT and NYTimes800k toward that of COCO, with our learned hypernymization approach closing the gap further than the manual approach (JSD=0.529 vs. 0.584 for WIT, and JSD=0.501 vs. 0.552 for NYTimes800k).

\subsection{How well do we extract object mentions from captions?}
\label{sec:results:mentions}
We report average precision=0.90 and recall=0.48 across classes on COCO train 2017 for the \verb+ExactMatch+ mention extraction method. We expected high precision since object names are generally used only to describe the objects they refer to, and we expected low recall because they are not the only way those objects are referred to. For example, a mention of ``person'' likely corresponds to a person in the image, but synonyms like ``woman'' are often used to describe an image with a person in it.

\subsection{What is the impact of hypernymization on grounding?}
\label{sec:results:grounding}
Table~\ref{table:results} reports our pretraining evaluation on WIT, NYTimes800k, and their hypernymized versions. Results on original versions of the datasets (Table~\ref{table:results:baselines}) are repeated for ease of comparison.
\begin{table}[htbp]
    \centering
    \begin{tabular}{l|c|c}
         \textbf{Dataset} & \textbf{All classes} &\textbf{\makecell[c]{High-precision and\\In DBPedia (7)}} \\
        \hline \hline
        COCO &  54.3 & 60.8  \\
        \hline \hline
        WIT & 38.5 & 54.6\\
        \hline
        \makecell[r]{Manual hypr.} & \textbf{38.9} & 52.9 \\
        \hline
        \makecell[r]{Learned hypr.} & 37.9 & \cellcolor{lightgray} \textbf{55.1} \\
        \hline \hline
        NYTimes800k & 40.5 & 55.8  \\
        \hline
        \makecell[r]{Manual hypr.} & \textbf{40.6} & 51.6 \\
        \hline
        \makecell[r]{Learned hypr.} & 40.3 & \cellcolor{lightgray}\textbf{56.2}
    \end{tabular}
    \caption{Evaluation results on COCO val 2017 (in percentage). Grounding mAP are reported  for all COCO classes and for those in the DBPedia Ontology and for which ExactMatch achieves precision higher than 0.9. Bold: highest performance per column per dataset. Shaded cells: results of note.}
    \label{table:results}
\end{table}

From Table~\ref{table:results} (middle column), manual hypernymization increases performance from mAP=38.5\% to 38.9\% for WIT, while we observe how our learned hypernymization approach overall underperforms the baseline (mAP=37.9\% v. 38.5\% for WIT, and 40.3\% v. 40.5\% for NYTimes800k).

When focusing on classes for which \verb+ExactMatch+ achieves precision $\ge 0.9$ and that are included in the DBPedia Ontology, our learned method boosts performance (from mAP=54.6\% to 55.1\% in WIT and from mAP=55.8\% to 56.2\% in NYTimes800k), while manual hypernymization does not. This confirms that the training data we created if effective in teaching a language model to perform hypernymization. We expect increasing the number of classes the language model can learn to hypernymize will translate into better overall grounding.

\subsection{What is the impact of hypernymization on open-vocabulary detection?}
\label{sec:results:ovr}
Table~\ref{table:results:ovr} reports mAP@0.5 for our open-vocabulary detection experiments on base classes only, target classes only, and both sets combined (``general''). 
\begin{table}[htbp]
    \centering
    \begin{tabular}{l|c|c|c|c|c}
         & \textbf{Base} & \textbf{Target} & \multicolumn{3}{c}{\textbf{Generalized}} \\
         \hline 
         & & & \textbf{Base} & \textbf{Target} & \textbf{All} \\
         \hline \hline
         COCO & 46.8 & 27.5 & 46.0 & 22.8 & 39.9 \\
         \hline \hline
         WIT &  43.8 & 6.6 & 43.2 & 3.9 & 32.9\\
         \hline
         \makecell[r]{Manual hypr.} & \textbf{44.0} & \cellcolor{lightgray}\textbf{7.9} & 43.3 & \cellcolor{lightgray}\textbf{4.4}& \textbf{33.1} \\
         \hline 
         \makecell[r]{Learned hypr.} & 43.9 & 3.8 & 43.4 & 1.2 & 32.4\\
         \hline \hline
         NYTimes800k & \textbf{43.1} & 6.7 & \textbf{42.4} & 3.2 & \textbf{32.1} \\
         \hline
         \makecell[r]{Manual hypr.} & 42.9 & 7.9 & 42.0 & \cellcolor{lightgray}\textbf{4.1} & 32.0\\
         \hline
         \makecell[r]{Learned hypr.} & 42.7 & \cellcolor{lightgray}\textbf{8.4} & 41.8 & \cellcolor{lightgray}\cellcolor{lightgray}\textbf{4.1} & 31.9
    \end{tabular}
    \caption{Open-vocabulary detection mAP@0.5 (in percentage). Results for COCO training are taken from~\cite{Zareian2021}. Bold: highest performance per column per dataset. Shaded cells: results of note.}
    \label{table:results:ovr}
\end{table}

Baseline models on WIT and NYTimes800k perform comparably. As expected, in both Wikipedia and news data, hypernymization is successful at improving performance, especially on target classes (mAP@0.5=7.9\% from 6.6\% for WIT and mAP@0.5=8.4\% from 6.7\% for NYTimes800k), although the manual hypernymization approach is more successful in WIT, while the learned one is in NYTimes800k. These results suggest that models pretrained on hypernymized captions may learn robust features that result in higher performance on target classes, for which no bounding-box supervision is available. On the other hand, for the base classes where supervision \emph{is} available, performance is similar across dataset versions, suggesting fine-tuning may dominate over any benefit from pretraining. In addition, the best hypernymization method may depend on each dataset's characteristics. On one hand, learned hypernymization could be better suited for datasets more similar to COCO (JSD between COCO and NYTimes800 0.557 vs. 0.597 for COCO and WIT from Table~\ref{table:results:jsd}), where our manipulations on COCO captions are able to better mimic the NE-rich data characteristics. On the other hand, the manual hypernymization method could be better suited for NE-rich data whose subtleties are harder to artificially reproduce. 

\section{Discussion and Conclusions}
\label{sec:discussion}
In this work, we studied the issue posed by the presence of named entities (NEs) in captions that naturally accompany images in domains like Wikipedia and news articles. We argued that NEs represent a missed learning opportunity when pretraining multi-modal models: if the caption mentioned the object by its category (e.g., ``person''), the model would better learn from image-caption pairs.

To address this problem, we introduced two ways to perform hypernymization: a manual approach based on NE recognition and DBPedia look-up, and a learned approach, where we trained a language model for hypernymization.

Our results show that models are able to ignore NEs during training, resulting in similar pretraining performance when NEs are left untouched and when they are removed (Table~\ref{table:results:baselines}). In addition, our analysis shows that both hypernymization strategies make NE-rich captions more similar to NE-free COCO captions (Tables~\ref{table:results:datasets_stats_wit},~\ref{table:results:datasets_stats_nyt}, and Table~\ref{table:results:jsd}) and that hypernymization can improve grounding (Table~\ref{table:results}). Finally, the benefit of hypernymization persists for open-vocabulary detection (Table~\ref{table:results:ovr}), especially on classes not seen during training.

This study has some limitations. For instance, the relative low number of COCO classes in the DBPedia Ontology limits the beneficial effect hypernymization can have on grounding-based pretraining. Our pretraining evaluation results restricted to such classes are encouraging, though; the more classes we can teach a language model to hypernymize, the more hypernymization could improve pretraining. 
Our pretraining evaluation approach also has some drawbacks. For instance, automatically extracting mentions of objects of interest is still imperfect despite its very high average precision. To limit the impact of erroneously extracted mentions, we focused our analysis on classes with very high precision only (Table~\ref{table:results}). 
Finally, we notice that improved pretraining performance does not always translate to improved downstream performance (Tables~\ref{table:results} and~\ref{table:results:ovr}). Other studies~\cite{ericsson2021well,kolesnikov2019revisiting} have started investigating the relationship between pretraining and downstream performance, which remains an active area of research.

\textbf{Societal impact:} We propose hypernymization as a way to better extract self-supervision from a dataset. For this reason, our method would further any type of bias present in the data, although the size of both WIT and NYTimes800k should reduce the likelihood of such biases by increasing diversity in the included data. In addition, the hypernymization process may also introduce bias if NEs for, say, people of a certain gender or race are more likely to be hypernymized. The use of a comprehensive resource like DBPedia in our hypernymization approaches ameliorates this issue since it makes it less likely to only include NEs for specific subgroups of people.

\textbf{Acknowledgments:} This material is based upon work supported by the National Science Foundation under Grant No. 2046853. GN was also supported by a University of Pittsburgh Intelligent Systems Program fellowship.

\bibliographystyle{ACM-Reference-Format}
\bibliography{main}


\end{document}